**Universidad Tecnológica del Uruguay, Rivera, Uruguay**

# RECONOCIMIENTO DE OBJETOS A PARTIR DE NUBE DE PUNTOS EN UN VEHÍCULO AÉREO NO TRIPULADO

Rodriguez, Anthony, anthony.rodriguez@estudiante.utec.edu.uy[1]
de Freitas Vidal, Marion Agustina, marion.defreitas@estudiantes.utec.edu.uy[2]
Suarez, Richard, richard.suarez@estudiantes.utec.edu.uy[3]

Grando Bedin, Ricardo, ricardo.bedin@utec.edu.uy[3]
Da Silva Kelbouscas, André, andre.dasilva@utec.edu.uy[4]

[1]Universidad Tecnológica del Uruguay, Rivera, Uruguay
[2]Universidad Tecnológica del Uruguay, Rivera, Uruguay

[3]Universidad Tecnológica del Uruguay, Rivera, Uruguay
[4]Universidad Tecnológica del Uruguay, Rivera, Uruguay

**RESUMEN**

En la actualidad la investigación en robótica, inteligencia artificial y drones están avanzando a nivel exponencial, se relacionan de forma directa o indirecta con varias áreas de la economía, desde agricultura hasta la industria. Con ese contexto, este proyecto abarca dichas temáticas guiandolas, buscando proveer un framework que sea capaz de ayudar a desarrollar nuevos futuros investigadores. Para tal, utilizamos un vehículo aéreo que funciona de forma autónoma y que es capaz de hacer un mapeo del escenario y proveer informaciónes útiles al usuario final.

Esto ocurre a partir de una comunicación entre un lenguaje de programación sencillo (Scratch) y uno de los sistemas operativos de robots más importantes y eficientes de la actualidad ROS (Robot Operating System). Así logramos desarrollar una herramienta capaz de generar un mapa 3D y detectar objetos utilizando la cámara acoplada al dron. Si bien esta herramienta puede ser utilizada en los campos avanzados de la industria también es un progreso importante para el sector de la investigación. Se aspira a la implementación de esta herramienta en instituciones de nivel intermedio para brindar la capacidad de realizar proyectos de alto nivel a partir de un lenguaje de programación simple.

*Palabras claves:* *Scratch, ROS, Orb Slam.*

**ABSTRACT**

Currently, research in robotics, artificial intelligence and drones are advancing exponentially, they are directly or indirectly related to various areas of the economy, from agriculture to industry. With this context, this project covers these topics guiding them, seeking to provide a framework that is capable of helping to develop new future researchers. For this, we use an aerial vehicle that works autonomously and is capable of mapping the scenario and providing useful information to the end user.

This occurs from a communication between a simple programming language (Scratch) and one of the most important and efficient robot operating systems today (ROS). This is how we managed to develop a tool capable of generating a 3D map and detecting objects using the camera attached to the drone. Although this tool can be used in the advanced fields of industry, it is also an important advance for the research sector. The implementation of this tool in intermediate-level institutions is aspired to provide the ability to carry out high-level projects from a simple programming language.

## 1 - INTRODUCCIÓN

El concepto SLAM (Simultaneous localization and mapping) plantea la problemática de un robot que explora un entorno desconocido. Cuando el robot explora el entorno, este tiene como objetivo obtener un mapa del mismo y a su vez ubicarse utilizando este. SLAM es el uso de modelados de ambientes detallados o la ubicación precisa de un robot, concepto este que fue una de las bases del proyecto. Para llevar a cabo este trabajo se utilizó un sistema operativo de robots llamado ROS que es un marco de código abierto para darle funcionalidades a los robots. ROS sirve como una plataforma de código común y utilización de distintos tipos de lenguajes de programación que permite compartir códigos e ideas de forma sencilla, pudiendo así lograr con mayor rapidez la movilidad de los robots.

Este proyecto fue incentivo para participar del Hackathon de Drones Uruguay 2022, competencia que tenía como pauta inicial la programación de un dron Tello a través de programación en bloques. La programación en bloques entretanto es limitada por ser una langage de introducción sin tener soporte a sistemas más avanzados como de inteligencia artificial y visión computacional, una vez que esas temáticas demandan complejidad de programación que sólo los lenguajes Python y C++ tienen.

De esta forma, para cumplir con las pautas de la competencia fue utilizado el lenguaje de programación Scratch. Scratch es según sus autores es un medio para que jóvenes puedan expresar sus ideas y brindar una herramienta que de una forma creativa facilite el uso de los ordenadores (Resnick, M., et al., 2003). Una vez desarrollado con Scratch, con el objeto de lograr que el dron DJI Tello pueda realizar un mapeo y ubicarse en el espacio, se utilizó una extensión entre Scratch 3 y ROS ya existente que fue creada por japoneses. A su vez se realizó la comunicación entre ROS y Orb Slam, un software que cumple con el concepto SLAM. Ese proyecto logró el segundo lugar en el Hackathon de Drones Uruguay 2022.

Nuestra herramienta logra diseñar un mapa utilizando una nube de puntos 3D, es el paso previo a la creación de un modelo preciso del mundo real. Es el punto de partida de una realidad digital, un mapa de puntos en el espacio que se procesan para convertirse en modelos 3D de casi cualquier objeto. A gran escala, esto incluye edificios, fábricas, plantas de fabricación, escenas de crímenes/accidentes, infraestructuras civiles, lugares históricos, paisajes urbanos y mucho más. Los modelos 3D basados en datos de nubes de puntos se utilizan en un número creciente de industrias para la visualización, la planificación y la personalización.

## 2 - METODOLOGÍA

Este proyecto está enfocado en la programación de drones utilizando Scratch, ROS y Orb Slam. El mismo se puede explicar en tres etapas distintas (Comunicación Scratch-ROS, Comunicación ROS-OrbSLAM, Sistema de mapeo). Cada etapa tiene un propósito trascendente a la implementación de problemáticas reales, el desarrollo de la etapa inicial parte de un dron comercial específico (DJI Tello, Figura 1), este es un dron programable desarrollado con fines educativos. En la figura 2 podemos observar las características de dicho drone.

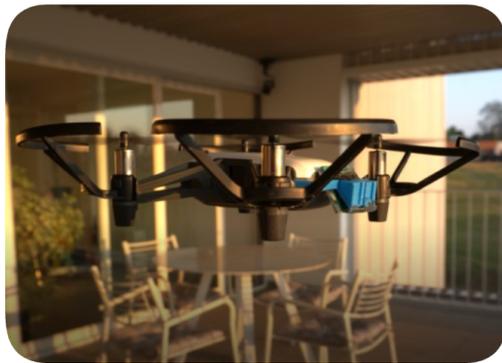

**Características Drone Tello**

| | |
|---|---|
| Peso | 80 g |
| Dimensiones | 9.8 x 9.2 x 4,1 cm |
| Duración de batería | 13 min |
| Distancia control remoto | 150 metros |
| Altura máxima de vuelo | 50 metros |
| Velocidad máxima de vuelo | 30 km/h |
| Cámara | 720p HD |
| Batería | 3.8V 1.100mAh |
| Otros | Modo 8D Flips y acrobacias 360º<br>Estabilización electrónica de imagen |

Figura 1. Ejemplo del drone utilizado (DJI TELLO).   Figura 2. Características del drone utilizado (DJI TELLO).

### 2.1 - EXTENSIÓN SCRATCH 3 Y ROS

Para la programación de nuestro dron se utilizó la extensión de Scratch 3 como herramienta por la libertad que brinda al permitir la comunicación con interfaces externas, siendo en este caso una comunicación establecida con plataformas habilitadas para el sistema operativo de robots (ROS). Dicha extensión funciona a través de la integración de bloques de utilidades para crear y manipular objetos de tipo JSON, formato de texto sencillo para el intercambio de datos, para la transmisión de datos, estos se integran a las variables de Scratch y se utilizan para la representación de mensajes ROS.

Esta extensión que conecta el lenguaje de programación visual de Scratch con ROS fue desarrollada por estudiantes de la universidad de Tokio con el objetivo de habilitar un rápido desarrollo en la creación de prototipos de interfaces destinadas a la programación de robots, teniendo como resultado final un diseño del algoritmo práctico, sencillo y confiable para la ejecución de tareas. Nuestro equipo se apoyó en esta herramienta debido a la participación del proyecto en la Hackathon de Drones Uruguay 2022, para la programación de drones fue requisito utilizar la programación visual de Scratch como soporte, lo cual en adición a la extensión implementada nos aportó herramientas que utilizamos para desarrollar lo que ideamos como una solución a nuestra problemática.

Figura 3. (Código Scratch con Interface con ROS)

En la figura 3 podemos observar el código utilizado para darle una trayectoria al dron y enviar la información necesaria a los tópicos que están siendo ejecutados dentro de ROS, en los tres primeros bloques estamos asignando valores a distintas variables que luego van a ser publicadas dentro de dichos tópicos, en otras palabras este código está dando la señal al dron para moverse sobre el espacio, este movimiento es directamente proporcional a la nube de puntos generada por el software Orb-SLAM como se puede observar en las figuras 4 y 5.

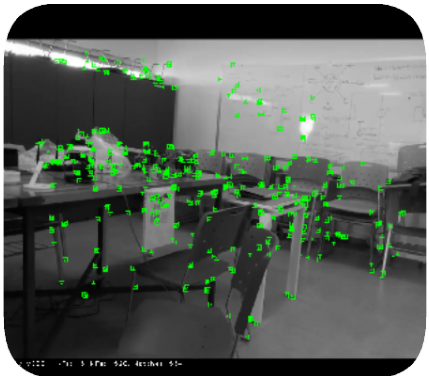
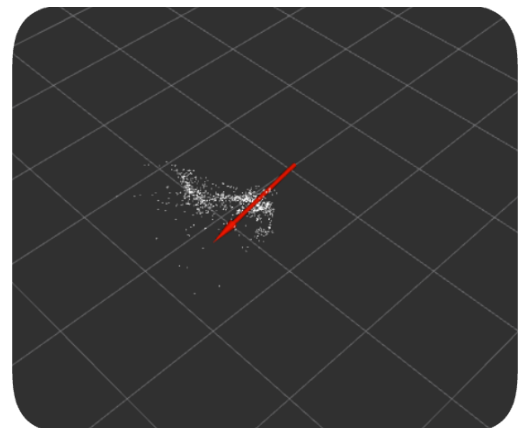

Figura 4. (Câmara del Dron desde Orb-Slam).                    Figura 5. (Nube de Puntos desde Orb-Slam).

**2.2 - COMUNICACIÓN ENTRE ORB SLAM Y ROS**

Para lograr entender la importancia de la comunicación entre Orb SLAM y ROS es necesario saber las capacidades y funcionalidades de tal. Para esta instancia utilizamos un software llamado ORB-SLAM2 (figura 6, Ejemplo). ORB-SLAM2 es un sistema SLAM completo para cámaras monoculares, estéreo y RGB-D, que incluye capacidades de reutilización de mapas, cierre de bucle y relocalización. El sistema funciona en tiempo real en CPU estándar en una amplia variedad de entornos, desde pequeñas secuencias portátiles en interiores hasta drones que vuelan en entornos industriales y automóviles que circulan por una ciudad. Nuestro back-end basado en el ajuste de haz con observaciones monoculares y estéreo permite una estimación precisa de la trayectoria con escala métrica. Este sistema incluye un modo de localización ligero que aprovecha las pistas de odometría visual para regiones no mapeadas y coincidencias con puntos del mapa que permiten una localización de desviación cero.

Dicha comunicación nos permite recibir información del software ORB-SLAM2 y enviar al sistema operativo de robot (ROS) para así lograr manipular con libertad el mapa ya generado, en la figura 6 podemos observar un ejemplo.

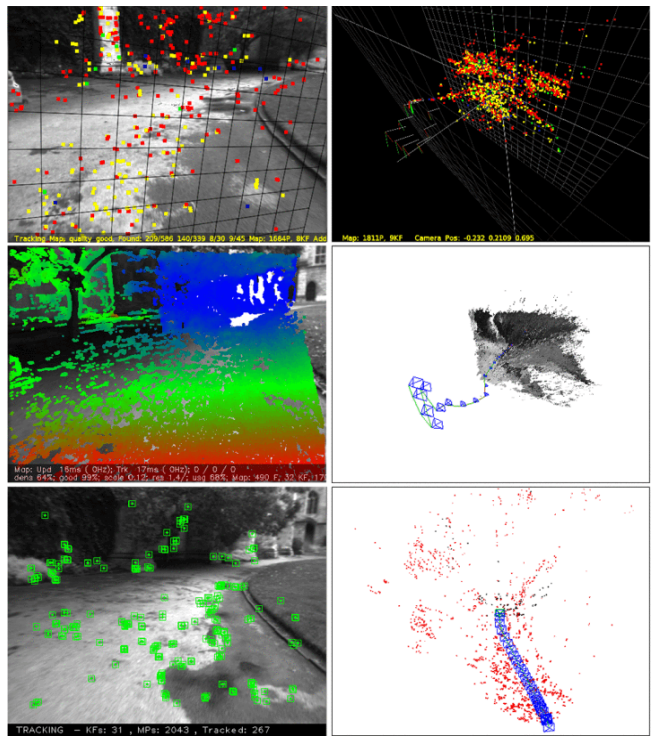

Figura 6. Ejemplo

### 2.3 - SISTEMA DE MAPEO

En esta etapa a partir del conjunto de comunicaciones Scratch, ROS y Orb-Slam2 se desarrolló una aplicación especifica, se logró construir una herramienta capaz de generar a través de la cámara de un dron un mapa 3d del espacio donde está sobrevolando, este sistema puede ser utilizada en diferentes ámbitos de la industria como por ejemplo minería, arquitectura, topografía de terrenos e incluso en seguridad. El dron puede recorrer una trayectoria definida por la programación en Scratch y la comunicación con ROS nos permite publicar directamente a los tópicos en ejecución y a su vez recibir información del software Orb-Slam2, debido a esto podemos manipular el dron y paralelamente a esto nuestra herramienta puede ir generando el mapa 3D.

### 3 - PRUEBAS PRÁCTICAS

Fueron realizados dos ensayos prácticos en el Hackathon de Drones Uruguay 2022, realizado en Mercedes, Uruguay. En el primer test el dron realizó el mapeo del ambiente y en el segundo test el dron hizo una navegación entre aros para testar la capacidad de navegación. Figura 7 muestra el dron realizando la tarea y un video (https://youtube.com/shorts/Hs8kPCLeRTE?feature=share) fue grabado mostrando en vehículo realizando las tareas.

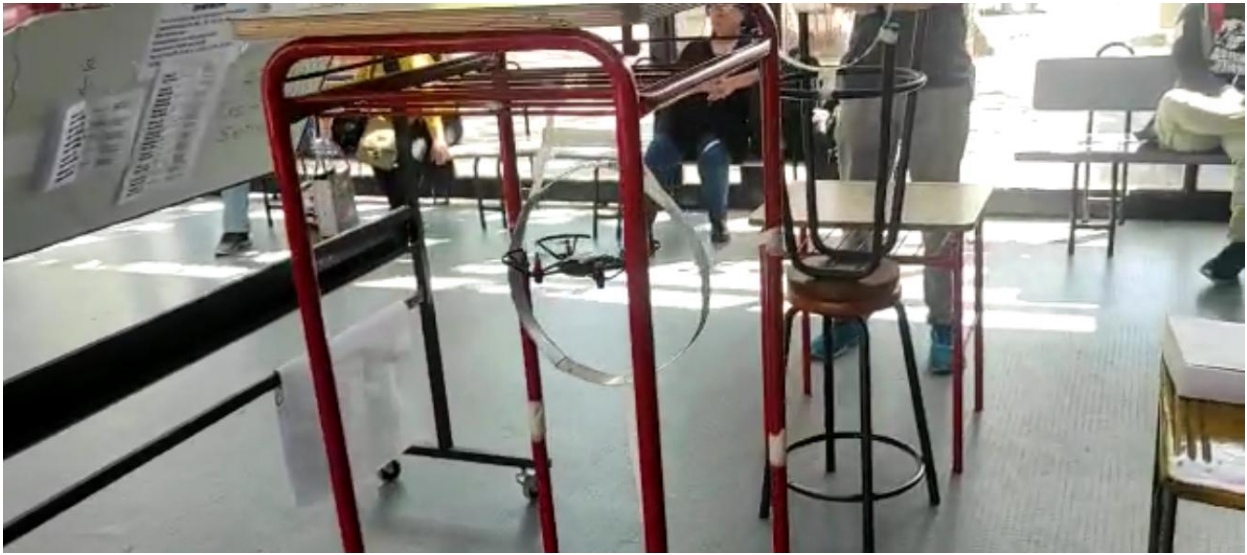

Figura 7. Dron realizando testes práticos

## 4 - CONCLUSIÓN

Se considera que haber logrado la comunicación entre Scratch, ROS y Orb Slam, genera un gran avance para el área de investigación para todo entorno académico ya que es una herramienta que facilita el acceso a gran cantidad de funciones que tiene ROS utilizando un lenguaje de programación simple tal como Scratch. Se utilizó el DJI Tello porque forma parte de las bases y condiciones del Hackaton, sin embargo este proyecto puede ser implementado en cualquier otro dron que tenga integrada una cámara.

Una de las limitaciones que presenta el DJI Tello es su baja autonomía, este tiene una duración aproximada de siete minutos, con lo cual es difícil lograr un mapeo completo. Además, para perfeccionar el mapa de puntos se considera implementar una red 5G para mejorar la calidad del video y a su vez enviaría la data de video de forma más rápida. Finalmente, con el objetivo de darle utilidad en el sector educativo, se creará la creación de una interfaz Scratch 3-ROS para aprendizaje en instituciones de educación media. Para progresar en el proyecto se planea la utilización de una gran cantidad de drones que trabajando de forma sincronizada puedan hacer el proceso del mapeo de forma más rápida y eficiente.

10**5 - REFERENCIAS**